\documentclass[final]{clv2025}
\jvol{vv}
\jnum{nn}
\jyear{2025}
\usepackage{setspace}

\usepackage{array}
\usepackage{booktabs}
\usepackage{multirow}
\usepackage{multicol}
\usepackage{amsmath,amssymb,amsfonts}
\usepackage{verbatim}
\usepackage{enumitem}
\usepackage{wrapfig}

\dochead{Last Words}

\runningtitle{Socially Aware Language Technologies}

\runningauthor{Yang, Hovy, Jurgens, Plank}

\pageonefooter{Action editor: Kevin Duh. Submission received: 30 August 2024; revised version received: 22 December 2024; accepted for publication: 6 February 2025.}
\begin{document}

\title{Socially Aware Language Technologies: Perspectives and Practices}

\author{Diyi Yang$^{\diamondsuit}$\thanks{$\diamondsuit$: Stanford University; $\clubsuit$: Bocconi University; $\spadesuit$: University of Michigan; $\heartsuit$: LMU Munich
}, Dirk Hovy$^{\clubsuit}$, David Jurgens$^{\spadesuit}$, Barbara Plank$^{\heartsuit}$}

\maketitle

\begin{abstract}
Language technologies have advanced substantially, particularly with the introduction of large language models. However, these advancements can exacerbate several issues that models have traditionally faced, including bias, evaluation, and risk. In this perspective paper, we argue that many of these issues share a common core: a lack of awareness of the social factors, interactions, and implications of the social environment in which NLP operates. We call this \textbf{social awareness}. While NLP is improving at addressing linguistic issues, there has been relatively limited progress in incorporating social awareness into models to work in all situations for all users. Integrating social awareness into NLP will improve the naturalness, usefulness, and safety of applications while also opening up new applications. 
Today, we are only at the start of a new, important era in the field.
\end{abstract}

\vspace{-0.08in}
\section{Introduction}\vspace{-0.08in}
Natural language processing (NLP) has made significant advances in recent years, thanks in part to the introduction of large pretrained language models (LLMs) based on Transformers \cite{brown2020language}.  As a result, performance on various NLP tasks has significantly improved, including machine translation, sentiment analysis, and conversational agents, to name but a few. NLP models appear to perform these tasks as well as, if not better than, humans \cite{tedeschi-etal-2023-whats}. On the other hand, an increasing number of problems and flaws with these models have been identified, which mean that NLP is working unevenly across users and situations. Some of these issues include bias~\cite{bolukbasi2016man,vida2024decoding}, toxicity~\cite{gehman2020realtoxicityprompts}, trust \cite{litschko2023establishing}, and fairness \cite{hovy-spruit-2016-social, blodgett-etal-2020-language, shah-etal-2020-predictive, elsherief2021latent}. For example, even basic components such as word embeddings, representing words in a mathematical space, can inadvertently capture and reinforce biases in training data, perpetuating stereotypes and inequalities \cite{bolukbasi2016man,gonen-goldberg-2019-lipstick,ryan-etal-2024-unintended}. Machine translation systems can produce translations with unintended biases or inaccuracies \cite{vanmassenhove-etal-2018-getting,hovy2020you}, potentially exacerbating cultural and societal misunderstandings \cite{bird-yibarbuk-2024-centering}. These concerns are exacerbated in widely used models such as LLMs~\cite[e.g.][]{vida2024decoding,kantharuban2024stereotype,wilson2024gender}.
All these aspects apply not only to English but also to the 7,000 languages available \cite{joshi-etal-2020-state}, adding complexity to the problem. 
Consequently, NLP ``works'' only for a subset of situations and people that use language technology \cite{held2023material}. 

We argue that many of these issues confronting modern NLP have a common core. They are caused by a failure to consider language (technologies) in a social context, i.e., in relation to social environments. We refer to these issues as \textbf{social awareness}, which refers to a system's awareness of social factors, contexts, and dynamics, as well as their implications for the broader social environment. Social awareness in models is currently undervalued. Traditional NLP models prioritized syntax, grammar, and lexicon, and their modern counterparts are direct descendants. However, they have not significantly progressed in understanding the sociocultural context and social interactions. In the linguistic terms of de Saussure, NLP has been mainly concerned with the abstract patterns of language (langue) without paying much attention to the concrete individual use of it (parole).
Operationalizing and integrating these complexities into today's LLMs is a significant challenge. However, we argue that addressing this issue is necessary to advance NLP. For example, the simple act of turn-taking in a dialogue, i.e., knowing whether, when, and how to respond requires a certain level of social awareness that LLMs currently lack \cite{ivey2024real}. Without it, conversations can be stilted and rude. 

Social awareness is ultimately integral to all modalities of AI, not just NLP, e.g., vision \cite{fathi2012social} and robotics \cite{breazeal2003emotion}, to name but a few examples. Social awareness governs the dynamics of human-human and human-AI interactions. Language is an essential tool in these processes for people to achieve a wide range of goals. NLP's potential insights and applications will inevitably be limited if it does not consider individual interactions, the context in which language is spoken, and the specific goals it should achieve. Knowing such goals or capabilities allows users to gain more trust in NLP systems \cite{litschko2023establishing}. 

Language is deeply intertwined with human society and culture, making it much more than just words and grammar. By modeling the social factors that influence language use, our models can broaden their scope and depth by better understanding and connecting with people \cite{hovy2021importance,hershcovich-etal-2022-challenges,pawar2024surveyculturalawarenesslanguage}.
The goal of this position paper is not simply to propose novel research directions, but to offer a comprehensive discussion of perspectives and practices about socially aware language technologies. It is intended for NLP researchers and practitioners interested in the societal impacts of language technologies, as well as HCI scholars and social scientists studying their influence on humans and society. 

\vspace{-0.15in}
\section{What Are Socially Aware Language Technologies?}\label{sec:social}\vspace{-0.08in}

We define \textbf{socially aware language technologies} as \textit{the study and development of language technologies from a social perspective}, allowing NLP systems to understand and respond to social signals expressed in language and broader physical and social environments.   
Socially aware systems can recognize social aspects and process socially driven meanings and implications behind language in the same way that humans do. 
In other words, a socially aware system exhibits emotional intelligence, cultural competence, and perspective-taking abilities, as discussed next, to ensure that advances in NLP are technologically sound and socially conscious.

\begin{wrapfigure}{r}{0.38\textwidth}
 \begin{center}
 \includegraphics[width=0.38\textwidth]{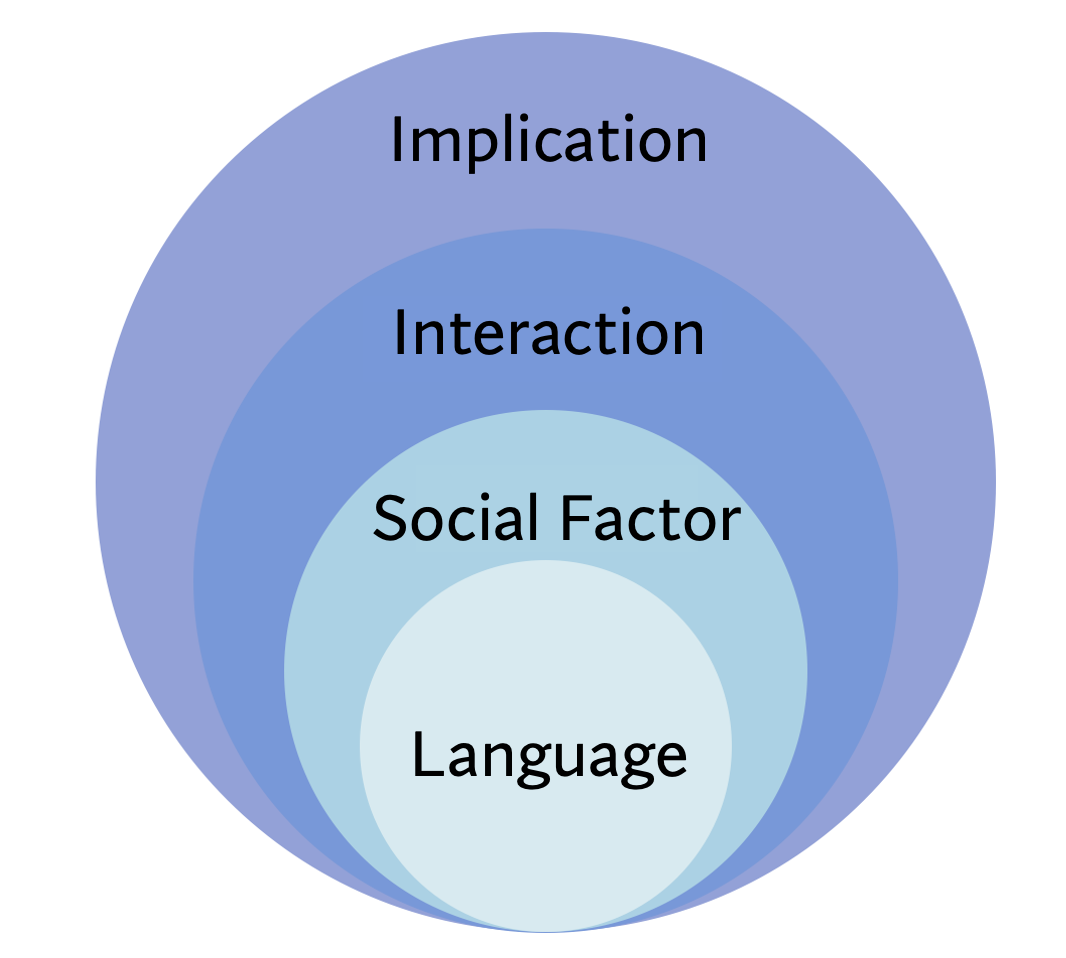}\vspace{-0.1in}
 \caption{Conceptual structure of socially aware language technologies: social factors, interaction, and implication. This is \textbf{not an exclusive partition}, but one way to understand the scope of social awareness. }\vspace{-0.05in}
 \label{fig:layers}
 \end{center}
\end{wrapfigure}

Prior work such as  \citet{pentland2005socially} defines \emph{socially aware computation} as systems that understand social signaling and context, and further argues that focusing on such dimensions can enhance collective decision-making and keep users informed.
The psychologist Daniel Goleman defines \emph{emotional intelligence} as four subsets: self-awareness, self-management, social awareness, and relationship management \cite{hernez2012social}. 
Emotional intelligence requires understanding and empathizing with others' emotions, which is related to Theory of Mind \cite{tomasello2014natural,premack1978does}. 
The emphasis on \textbf{social awareness} in NLP means creating tasks, models, and evaluations that consider social factors \citep{hovy2021importance} first (as illustrated in the inner two circles in Figure~\ref{fig:layers}), and then go further by including the full social context and the social dynamics communicated through language. 
Researchers and practitioners need to become aware of these social aspects to design socially aware NLP systems. 
Note that while social awareness includes aspects of cultural and personal identity, it does not require us to take a moral stance, and we do not prescribe \textit{what} perspectives models will have to take (and likely they will differ substantially among different countries and languages). In complex contexts where diverse values and priorities can lead to conflicting outcomes \cite{sorensen2024roadmap}, what benefits one group may harm another. With socially aware language technologies, we can gain a better understanding of such complex situations, as it advocates incorporating context and social factors such as culture into the design process, so that resulting technologies reflect the diverse viewpoints rather than prompting dominant ones.

We argue that developing socially aware language technologies must prioritize three key aspects: \textbf{social factors} (Section \ref{sec:soc_factors}), \textbf{interaction} (Section \ref{sec:soc_interaction}), and \textbf{implication} (Section \ref{sec:soc_implication}).
Figure \ref{fig:layers} shows that 
social awareness is often present near the design of the tasks and algorithms (social factors). Social awareness also plays a central role in the middle-ground of interactions and activities that humans have with NLP systems (interaction)  and the outer layer of impact (implication) that NLP systems may have on people and society.  By putting a strong emphasis on social factors, interaction, and implications, we hope that socially aware language technologies can facilitate better communication and align with human preferences and societal values. 

\vspace{-0.18in}
\subsection{Social Factors}
\label{sec:soc_factors}\vspace{-0.1in}
We define \textbf{social factors} as a wide range of social aspects that shape the way we understand language use, including but not limited to who is the speaker, who is the receiver, what is their social relation, in what context, guided by what kinds of social norms, culture, and ideology, and for what communicative goals \cite[see][]{hovy2021importance}. One can also greatly enrich such a list by incorporating insights and understandings around social factors from psychology, sociology, and other disciplines.
There is an increasing trend in doing so in the NLP community, and recent work has highlighted the importance and complexity of evaluating social attitudes, opinions, and values embedded within LLMs~\cite{ma-etal-2024-potential}.
Social factors, in particular, can motivate socially informed tasks by adding objective functions and tasks. Operationalizing social phenomena will expand the current pool of tasks to reflect users' needs better, resulting in increased user trust. Social knowledge can enhance existing representations in models \citep{nguyen2021learning} and impact opinions and
steerability of LLMs~\cite{wright-etal-2024-llm}. Social signals may offer alternative supervision for representation learning and next-word prediction. Current models have internal representations of social factors but do not seem to actively draw on them \citep{lauscher2022socioprobe}. With social awareness integrated into the pipeline, the outcome can have a social impact, not just on the typical task evaluation metric but also on people.

\vspace{-0.15in}
\subsection{Interaction}
\label{sec:soc_interaction}\vspace{-0.15in}
\textbf{Interaction} refers to the social exchanges and activities between humans and NLP systems, including relational, organizational, and cultural norms that govern interpersonal communication, as well as the evolving contexts in which these systems operate. 
Social science theories, such as those based on social influence and social norms, define critical dimensions of social interaction, providing insight into how humans interact and behave.
Such perspective posits that language is not an isolated construct but emerges as a product of social exchange and communication, aligning closely with the interactionism paradigm in sociology \cite{snyder1985personality}. Social norms govern social behavior and are defined as groups' shared standards of acceptable behavior. 
Integrating social norms into language adds expressivity beyond vocabulary and grammar. The work of \citet{lapinski2005explication} highlights the nuanced interplay between social norms and language, demonstrating that linguistic expressions frequently serve as vehicles for the expression and reinforcement of these norms.
These provide a rich and multifaceted foundation for our explorations of the complicated space of \textbf{social interaction}, including social exchange between individuals, other people in the context, and other activities surrounding the context. 
Language use, like self-perception \cite{cooley1902looking}, is influenced by others' perceptions of the language, especially as people interact with LLMs. 
Socio-technical NLP systems are part of a social interaction ecosystem in which users, developers, and stakeholders collaborate to develop, deploy, and use these technologies. 
Many factors and social phenomena are revealed in social interactions, such as 
power dynamics \cite{prabhakaran-etal-2013-upper}, trust \cite{litschko2023establishing} and user expectations~\cite{dhuliawala-etal-2023-diachronic}. 
NLP system design must consider how social interactions shape user experiences \cite{jakesch2023co,liu2022will} and impact the technology's adoption and effectiveness.

\vspace{-0.1in}
\subsection{Implication}
\label{sec:soc_implication}\vspace{-0.1in}

\textbf{Implication} refers to the impact of an NLP system on society, including both positive and negative effects.
Understanding the social implications of NLP is crucial for responsible development and sustainable use. This process involves assessing biases and stereotypes \citep{dev2022measures}, considering how systems affect global populations, not just those in the North \cite{song2023globalbench,ranathunga2022some}, investigating misinformation and dual use of NLP systems, examining concerns about job displacement \cite{eloundou2023gpts} and human-LLM alignment on other factors such as for example creativity~\cite{spangher-etal-2024-llms} and storytelling~\cite{tian-etal-2024-large-language}. 
Understanding social implications can also inspire model design, such as developing models that consider the implications of their outputs. For example, work in prompt safety can be viewed as an initial step towards imbuing models with a sense of what responses have harmful social implications 
\citep[e.g.,][]{bianchi2023safety}. As a society, we first have to understand these social implications to fully use NLP and mitigate its harmful effects.

\vspace{-0.1in}
\section{Situating Socially Aware Language Technologies}\vspace{-0.1in}

As with many other things, social awareness is best identified by its absence. 
Without social awareness, NLP technology will disregard social or cultural taboos, fail to consider personalized aspects of language applications, use language that the target audience cannot understand (due to age, education level, or other factors), or respond inappropriately or hurtfully (e.g., telling a suicidal user to kill themselves \cite{dinan-etal-2022-safetykit}) or in non-natural ways. 
Socially aware language technologies are related to many emerging topics, and as more work is being done or needs to be done around social and language technologies, there is a crucial need to differentiate socially aware language technologies from other approaches. \smallskip

\noindent
\textit{Differentiation} Socially aware NLP differs from \emph{personalization} because it aims to incorporate a broader context of language use, such as larger social and cultural groups. In contrast, personalization focuses more on the individual for a customized user experience~\cite[e.g.][]{flek-2020-returning}.
The concepts of socially aware NLP and ''\emph{NLP in a social context}'' are related but not the same.
Using NLP techniques to analyze and understand language use in social settings such as online communities, political discourse, and public opinion is called NLP in a social context or computational sociolinguistics~\cite{nguyen-etal-2016-survey}, sometimes through the text lens of Computational Social Science (CSS). CSS often develops NLP models to uncover patterns and trends in text to answer questions in the social sciences.
Another related concept is the \emph{theory of mind} \cite[ToM;][]{grant2017can,le2019revisiting,sap-etal-2022-neural}, which refers to the ability of models to reason about the mental state of others (e.g., intents, emotions, or beliefs). While ToM and socially aware NLP aim to improve models' ability to interact with humans more easily and socially appropriately, ToM differs from socially aware NLP by focusing on 
inferring others' mental states through attributing mental states and understanding intentions.
\emph{Human-Centered NLP} and socially aware NLP both emphasize how to make NLP aware of human factors and align with real-world needs, including ethical considerations and inclusiveness of languages and cultures~\cite{hucllm-2024-human}. 
Human-centered NLP, on the other hand, focuses on user-centered design to create systems tailored to user needs and is frequently based on iterative design, usability testing, and human-in-the-loop approaches to improve human-system interactions. 
Similar comparisons apply to the difference between \emph{human-like} and \emph{social awareness}. Human-like AI aims to mimic human-to-human interactions by making them natural and familiar to humans. At the same time, social awareness \emph{further} encourages appropriate and considerate interactions with the social environment. 

\vspace{-0.1in}
\section{Building Socially Aware Language Technologies}\vspace{-0.1in}

While NLP has recognized the importance of social language and begun to develop models capable of interacting socially, there are still significant gaps. Here, we motivate the key considerations and strategic goals for closing these gaps.

\vspace{-0.1in}
\subsection{Considerations for Socially Aware NLP}\label{sec:consideration}
\vspace{-0.1in}

Just as language grounding models benefit from diverse imagery to map to language, socially aware models need exposure to massive amounts of diversity---persons, values, relationships, etc.---in order to learn effective representations of social information and learn to reason about it. Humans naturally form categories to describe stereotypical social entities (e.g., persons, groups, relationships), which help guide expectations for behavior and communication in the absence of more explicit information \citep{blair2001imagining,rhodes2019development}; while crude and potentially harmful, such stereotypes can serve as effective priors that simplify social processing \textit{but} are only an initial starting point in a representation of others and one that is updated through interaction---bringing a richness and nuance to how we think of and relate to each other. Given the complexity of social awareness, we argue that models need sufficient diversity to effectively learn these categories so that they can easily adapt and reason about the social cues seen in different settings \cite{yin2021including,sharma2023cognitive,wang-etal-2024-bridging}.
Thus, we argue for the following considerations: social awareness and social reasoning must (C1) be grounded in socially diverse data, with fluid representations that move from the categorical to the bespoke; (C2) extend to interactive contexts, diverse cultural settings, and non-text modalities, (C3) be adaptable to different context and be capable of adapting during interaction, and (C4) be interpretable or explainable in how social awareness is used to reason.

\vspace{-0.22in}
\subsection{LLMs are Not Socially Aware Language Technologies Yet}\label{sec:task}\vspace{-0.1in}

The advanced language capabilities of LLMs have opened up many new interactive and seemingly social applications. However, we argue that LLMs are not yet socially aware and that we need new goals and measurements to gauge progress and move beyond traditional NLP benchmarks for social tasks \citep[e.g.,][]{choi2023llms,ziems2023can}.

(1) \textit{Operationalization and measurement of social awareness.}  Many recent studies have started to quantify social awareness \cite{rathje2024gpt} like measuring social relations \cite{iyyer2016feuding,choi2021more} or recognizing inappropriate content \cite{kumar2024watch}. LLMs are shown to struggle with these social signals \cite{ziems2023can}, calling for new algorithms and systems to deal with them. It becomes increasingly important to operationalize different aspects of social awareness based on theories and insights from social science in order to determine whether and to what extent LLMs have exhibited social awareness. Further, such evaluations must go beyond static benchmarks or multiple-choice questions to operate in an interactive way. 

(2) \textit{Behavioral expectations from social science theories}. Experimental work in the social sciences such as Psychology and Behavioral Economics has generated clear expectations for a variety of human behaviors, such as trust \cite{evans2009psychology} and risk aversion \cite{dohmen2005individual}, that depend on social awareness. When prompted with similar settings and information, these insights can serve as references for external behaviors that demonstrate LLMs are accurately reasoning about social awareness in a human-like manner \cite{Park2024GenerativeAS}. Such experimental measurement allows us to test for whether models recognize the social factors in play and to interact accordingly. 

(3) \textit{Inference of social context and use in reasoning}. Humans learn to recognize social cues as they mature and reason about this information \cite{thompson2007development,sher2014children}. While LLMs are increasingly capable of complex reasoning tasks \cite{huang2023towards}, they are still only beginning to learn to recognize social information through interaction \citep[e.g.,][]{zhang2023exploring} and to be able to explicitly incorporate and explain how this information influences their reasoning \cite{gandhi2024understanding}. As one example domain, many social games involve reasoning not only about the game state but players' mental states and the social implications of certain actions \cite{colman2003cooperation}. 
Games can provide new domains for assessing models' abilities to learn social factors from cues across turns and reason about other players. 

(4) \textit{Deployment of socially aware behavior in practical applications}. Technologies that use the social factors and implications in real-world applications provide rich ground for assessing progress in social awareness. Recent work has targeted applications such as therapy and coaching \cite{suh2024special}, inclusive technologies by providing access to services for people with disabilities \cite{guo2020toward}, and language technologies for positive impact \cite{jin2021good}. However, clear gaps exist between human social behavior in these applications and the capabilities of LLMs. 

(5) \textit{Understanding how socially aware language technologies affect people and society}. With the increase of LLM-empowered applications, it becomes critical to understand these broad implications, which include  
how LLM-empowered applications affect how people communicate and interact with each other \cite{liu2022will}, reinforce stereotypes or biases \cite{dev2022measures} and affect public trust, education, and the labor market \cite{eloundou2023gpts}, as well as how they inform policy and regulation. 

Taking all together, there is a critical need for a sub-field of ''\emph{socially aware language technologies}'' due to the increasing work on social and language technologies. Within this new subfield, we must ensure that language processing advances are technologically sophisticated and socially conscious. A unified subfield focused on this goal would allow researchers to systematically address the challenges of embedding social intelligence into language models, allowing for more precise communication among scientists, policymakers, and the general public. 
Recognizing ''\emph{socially aware language technologies}'' is a strategic step towards a future where language technology responsibly interacts with human society.

\vspace{-0.15in}
\section{Historical View of Socially Aware Language Technologies}\vspace{-0.1in}

Early AI was conceived in a much more holistic manner than the fragmented space that exists today. Its goal was to produce \textit{human-like} behavior, which required a tight coupling of different aspects and disciplines. That goal assumed social awareness, even if not explicitly stated \cite{turing1950,mccarthy2006proposal}. Moravec's paradox \cite{moravec1988mind}, often summarized pithily as, ''\emph{In AI, easy things are hard, and hard things are easy}'' \cite{pinker2003language}, has singled out \textit{social awareness} and \textit{motion} as the main areas where AI models have difficulty matching human performance even on simple tasks (while outperforming humans on tasks that require patience or logic). 
Over time, AI specialized into subfields, which shifted their focus to easier-to-solve tasks. Those were typically information or logic-based and did not require social awareness. As a result, NLP has spent a long time focusing on information-rich linguistic analysis tasks like parsing. Recent research has focused on language's social, cultural, and demographic aspects \cite{hovy2021importance,c3nlp-2023-cross}. 

LLMs' strong performance on various language understanding tasks may create the superficial impression that these models are now socially aware. However, many of the tasks they excel at are language-only problems that do not necessitate social awareness. Furthermore, tasks designed to demonstrate social, psychological, or emotional aspects of models frequently operate on a flawed premise. For example, \citet{sap-etal-2022-neural} demonstrated that while we can administer ToM tests to LLMs, the question itself is ill-posed. A human subject's ToM can be gauged via question-based psychological tests because their responses are influenced by their complex inner workings. In contrast, LLMs respond by generating a list of likely words. No ToM required. Similarly, \citet{shu2024you} demonstrates that while LLMs can generate answers to psychometric questionnaires like personality tests, their answers are inconsistent and lack awareness of the premise.
Even when human and model responses are similar, they stem from very different causes. In the absence of explicit modeling, it is unclear whether LLMs would develop social capabilities by themselves.

As we enter the era of LLM-dominated NLP, the next logical step is to tackle ``harder'' problems. Applying Moravec's paradox, the next more difficult area for NLP would be either motion (less applicable and addressed by robotics, but possible in the context of multimodality) or social awareness. This step aligns with a growing societal need. However, making progress in this area means answering difficult questions:
Is it possible to gain social awareness gradually and/or systematically? Can we teach our models how humans develop social awareness? 
Despite the difficulty of replicating human social awareness in machines, we advocate for the development of NLP systems capable of learning and recognizing social awareness over time, as well as responding to these cues in a more human-like manner.

\vspace{-0.1in}
\section{The Future of (Socially Aware) Language Technologies} \vspace{-0.08in}

NLP is not the first field to focus on the abstract over the concrete. Linguistics used to view language as separate from all other cognitive (and physical) abilities. While this abstract framing allowed for studying specific aspects in isolation and developing theories and models, it obscured the overall picture. Sociolinguistics, psycholinguistics, and other subfields have worked hard to reintroduce the importance of ``extraneous'' factors into the linguistic mainstream. 

Today, NLP follows a similar trajectory.
Many traditional NLP tasks have become obsolete as LLMs play a more significant role in AI research. However, as the power of those models grows, we are increasingly free to think about their use in a techno-social environment \cite{blodgett-etal-2020-language, tedeschi-etal-2023-whats, abercrombie-etal-2023-mirages}. 
With language models providing the foundation for natural language generation and analysis, we can (re)focus on the social aspects of language modeling.
Understanding the social aspects of language technologies requires a focus on emotional intelligence, cultural factors, values, norms, social interaction, and broader social implications. 
Developing socially aware NLP requires more than simply building models that recognize social factors, as \citet{hovy2021importance} have suggested; it also involves examining how these NLP systems interact with both social and physical environments, as well as their broad social implications. As long as NLP systems exist, social awareness will remain essential, because social factors, interactions, and their implications are integral to any human engagement with these technologies.

Socially aware NLP is likely to transform industries and societal functions while also shaping the broader field of AI, including audio, vision, and robotics, where social awareness can play an even more critical role. Integrating social awareness in robotics can enable the development of robots that can safely and effectively interact with humans (e.g., eldercare robots, service robots), and advancements in computer vision that enable systems to better interpret emotions, social interactions, and cultural contexts from visual data \cite{mittal2020emoticon,kwon2023toward, kruk2023impressions,achlioptas2021artemis}. 
In contrast, developing socially aware NLP can also introduce significant risks such as misunderstanding cultures, enforcing biases, and violating privacy. The misuse of such socially aware systems may also lead to over-reliance and echo chambers, as well as misinformation by bad actors. 
We should proceed with a keen awareness of ethics and risks \cite{barrett2023identifying}. 

In the future, we can look into how these models function as social agents, what social cues they read and understand, and what tasks requiring social awareness they can complete.
This pivot will necessitate new tasks, metrics, and approaches fundamentally different from the goals we have pursued as a field thus far. Most importantly, it will necessitate a re-alignment of the current fractured AI landscape: we will need to collaborate across disciplines to incorporate social awareness into our models. There are numerous unexplored research areas awaiting exploration.

Humans are more than language factories: language is only one component of our complex social interactions. We are not human because we speak---we speak because we are human. Language models, on the other hand, and at this point, are language factories capable of producing and processing words at astonishing rates but lacking the faculties that drive human language production and processing of world knowledge and social nuances. Socially aware language technologies can get us closer to AI's initial goals, advance the field, and help address many of the current issues we face.

\starttwocolumn
\bibliographystyle{compling}
\bibliography{ref}

\begin{thebibliography}{90}
\expandafter\ifx\csname natexlab\endcsname\relax\def\natexlab#1{#1}\fi

\bibitem[{Abercrombie et~al.(2023)Abercrombie, Curry, Dinkar, Rieser, and Talat}]{abercrombie-etal-2023-mirages}
Abercrombie, Gavin, Amanda Curry, Tanvi Dinkar, Verena Rieser, and Zeerak Talat. 2023.
\newblock Mirages. on anthropomorphism in dialogue systems.
\newblock In \emph{Proceedings of the 2023 Conference on Empirical Methods in Natural Language Processing}, pages 4776--4790, Association for Computational Linguistics, Singapore.

\bibitem[{Achlioptas et~al.(2021)Achlioptas, Ovsjanikov, Haydarov, Elhoseiny, and Guibas}]{achlioptas2021artemis}
Achlioptas, Panos, Maks Ovsjanikov, Kilichbek Haydarov, Mohamed Elhoseiny, and Leonidas~J Guibas. 2021.
\newblock Artemis: Affective language for visual art.
\newblock In \emph{Proceedings of the IEEE/CVF Conference on Computer Vision and Pattern Recognition}, pages 11569--11579.

\bibitem[{Barrett et~al.(2023)Barrett, Boyd, Bursztein, Carlini, Chen, Choi, Chowdhury, Christodorescu, Datta, Feizi et~al.}]{barrett2023identifying}
Barrett, Clark, Brad Boyd, Elie Bursztein, Nicholas Carlini, Brad Chen, Jihye Choi, Amrita~Roy Chowdhury, Mihai Christodorescu, Anupam Datta, Soheil Feizi, et~al. 2023.
\newblock Identifying and mitigating the security risks of generative ai.
\newblock \emph{Foundations and Trends{\textregistered} in Privacy and Security}, 6(1):1--52.

\bibitem[{Bianchi et~al.(2023)Bianchi, Suzgun, Attanasio, R{\"o}ttger, Jurafsky, Hashimoto, and Zou}]{bianchi2023safety}
Bianchi, Federico, Mirac Suzgun, Giuseppe Attanasio, Paul R{\"o}ttger, Dan Jurafsky, Tatsunori Hashimoto, and James Zou. 2023.
\newblock Safety-tuned llamas: Lessons from improving the safety of large language models that follow instructions.
\newblock \emph{arXiv preprint arXiv:2309.07875}.

\bibitem[{Bird and Yibarbuk(2024)}]{bird-yibarbuk-2024-centering}
Bird, Steven and Dean Yibarbuk. 2024.
\newblock Centering the speech community.
\newblock In \emph{Proceedings of the 18th Conference of the European Chapter of the Association for Computational Linguistics (Volume 1: Long Papers)}, pages 826--839, Association for Computational Linguistics, St. Julian{'}s, Malta.

\bibitem[{Blair, Ma, and Lenton(2001)}]{blair2001imagining}
Blair, Irene~V, Jennifer~E Ma, and Alison~P Lenton. 2001.
\newblock Imagining stereotypes away: the moderation of implicit stereotypes through mental imagery.
\newblock \emph{Journal of personality and social psychology}, 81(5):828.

\bibitem[{Blodgett et~al.(2020)Blodgett, Barocas, Daum{\'e}~III, and Wallach}]{blodgett-etal-2020-language}
Blodgett, Su~Lin, Solon Barocas, Hal Daum{\'e}~III, and Hanna Wallach. 2020.
\newblock Language (technology) is power: A critical survey of {``}bias{''} in {NLP}.
\newblock In \emph{Proceedings of the 58th Annual Meeting of the Association for Computational Linguistics}, pages 5454--5476, Association for Computational Linguistics, Online.

\bibitem[{Bolukbasi et~al.(2016)Bolukbasi, Chang, Zou, Saligrama, and Kalai}]{bolukbasi2016man}
Bolukbasi, Tolga, Kai-Wei Chang, James~Y Zou, Venkatesh Saligrama, and Adam~T Kalai. 2016.
\newblock Man is to computer programmer as woman is to homemaker? debiasing word embeddings.
\newblock \emph{Advances in neural information processing systems}, 29.

\bibitem[{Breazeal(2003)}]{breazeal2003emotion}
Breazeal, Cynthia. 2003.
\newblock Emotion and sociable humanoid robots.
\newblock \emph{International journal of human-computer studies}, 59(1-2):119--155.

\bibitem[{Brown et~al.(2020)Brown, Mann, Ryder, Subbiah, Kaplan, Dhariwal, Neelakantan, Shyam, Sastry, Askell et~al.}]{brown2020language}
Brown, Tom, Benjamin Mann, Nick Ryder, Melanie Subbiah, Jared~D Kaplan, Prafulla Dhariwal, Arvind Neelakantan, Pranav Shyam, Girish Sastry, Amanda Askell, et~al. 2020.
\newblock Language models are few-shot learners.
\newblock \emph{Advances in neural information processing systems}, 33:1877--1901.

\bibitem[{Choi et~al.(2021)Choi, Budak, Romero, and Jurgens}]{choi2021more}
Choi, Minje, Ceren Budak, Daniel~M Romero, and David Jurgens. 2021.
\newblock More than meets the tie: Examining the role of interpersonal relationships in social networks.
\newblock In \emph{Proceedings of the International AAAI Conference on Web and Social Media}, volume~15, pages 105--116.

\bibitem[{Choi et~al.(2023)Choi, Pei, Kumar, Shu, and Jurgens}]{choi2023llms}
Choi, Minje, Jiaxin Pei, Sagar Kumar, Chang Shu, and David Jurgens. 2023.
\newblock Do llms understand social knowledge? evaluating the sociability of large language models with socket benchmark.
\newblock In \emph{Proceedings of the 2023 Conference on Empirical Methods in Natural Language Processing}, pages 11370--11403.

\bibitem[{Colman(2003)}]{colman2003cooperation}
Colman, Andrew~M. 2003.
\newblock Cooperation, psychological game theory, and limitations of rationality in social interaction.
\newblock \emph{Behavioral and brain sciences}, 26(2):139--153.

\bibitem[{Cooley(1902)}]{cooley1902looking}
Cooley, Charles~Horton. 1902.
\newblock The looking-glass self.
\newblock \emph{The production of reality: Essays and readings on social interaction}, 6(1902):126--28.

\bibitem[{Dev et~al.(2023)Dev, Prabhakaran, Adelani, Hovy, and Benotti}]{c3nlp-2023-cross}
Dev, Sunipa, Vinodkumar Prabhakaran, David Adelani, Dirk Hovy, and Luciana Benotti, editors. 2023.
\newblock \emph{Proceedings of the First Workshop on Cross-Cultural Considerations in NLP (C3NLP)}. Association for Computational Linguistics, Dubrovnik, Croatia.

\bibitem[{Dev et~al.(2022)Dev, Sheng, Zhao, Amstutz, Sun, Hou, Sanseverino, Kim, Nishi, Peng et~al.}]{dev2022measures}
Dev, Sunipa, Emily Sheng, Jieyu Zhao, Aubrie Amstutz, Jiao Sun, Yu~Hou, Mattie Sanseverino, Jiin Kim, Akihiro Nishi, Nanyun Peng, et~al. 2022.
\newblock On measures of biases and harms in nlp.
\newblock In \emph{Findings of the Association for Computational Linguistics: AACL-IJCNLP 2022}, pages 246--267.

\bibitem[{Dhuliawala et~al.(2023)Dhuliawala, Zouhar, El-Assady, and Sachan}]{dhuliawala-etal-2023-diachronic}
Dhuliawala, Shehzaad, Vil{\'e}m Zouhar, Mennatallah El-Assady, and Mrinmaya Sachan. 2023.
\newblock A diachronic perspective on user trust in {AI} under uncertainty.
\newblock In \emph{Proceedings of the 2023 Conference on Empirical Methods in Natural Language Processing}, pages 5567--5580, Association for Computational Linguistics, Singapore.

\bibitem[{Dinan et~al.(2022)Dinan, Abercrombie, Bergman, Spruit, Hovy, Boureau, and Rieser}]{dinan-etal-2022-safetykit}
Dinan, Emily, Gavin Abercrombie, A.~Bergman, Shannon Spruit, Dirk Hovy, Y-Lan Boureau, and Verena Rieser. 2022.
\newblock {S}afety{K}it: First aid for measuring safety in open-domain conversational systems.
\newblock In \emph{Proceedings of the 60th Annual Meeting of the Association for Computational Linguistics (Volume 1: Long Papers)}, pages 4113--4133, Association for Computational Linguistics, Dublin, Ireland.

\bibitem[{Dohmen et~al.(2005)Dohmen, Falk, Huffman, Sunde, Schupp, and Wagner}]{dohmen2005individual}
Dohmen, Thomas, Armin Falk, David Huffman, Uwe Sunde, J{\"u}rgen Schupp, and Gert~G Wagner. 2005.
\newblock Individual risk attitudes: New evidence from a large, representative, experimentally-validated survey.
\newblock Technical report, DIW Discussion Papers.

\bibitem[{Eloundou et~al.(2023)Eloundou, Manning, Mishkin, and Rock}]{eloundou2023gpts}
Eloundou, Tyna, Sam Manning, Pamela Mishkin, and Daniel Rock. 2023.
\newblock Gpts are gpts: An early look at the labor market impact potential of large language models.
\newblock \emph{arXiv preprint arXiv:2303.10130}.

\bibitem[{ElSherief et~al.(2021)ElSherief, Ziems, Muchlinski, Anupindi, Seybolt, De~Choudhury, and Yang}]{elsherief2021latent}
ElSherief, Mai, Caleb Ziems, David Muchlinski, Vaishnavi Anupindi, Jordyn Seybolt, Munmun De~Choudhury, and Diyi Yang. 2021.
\newblock Latent hatred: A benchmark for understanding implicit hate speech.
\newblock In \emph{Proceedings of the 2021 Conference on Empirical Methods in Natural Language Processing}, pages 345--363.

\bibitem[{Evans and Krueger(2009)}]{evans2009psychology}
Evans, Anthony~M and Joachim~I Krueger. 2009.
\newblock The psychology (and economics) of trust.
\newblock \emph{Social and Personality Psychology Compass}, 3(6):1003--1017.

\bibitem[{Fathi, Hodgins, and Rehg(2012)}]{fathi2012social}
Fathi, Alircza, Jessica~K Hodgins, and James~M Rehg. 2012.
\newblock Social interactions: A first-person perspective.
\newblock In \emph{2012 IEEE Conference on Computer Vision and Pattern Recognition}, pages 1226--1233, IEEE.

\bibitem[{Flek(2020)}]{flek-2020-returning}
Flek, Lucie. 2020.
\newblock Returning the {N} to {NLP}: {T}owards contextually personalized classification models.
\newblock In \emph{Proceedings of the 58th Annual Meeting of the Association for Computational Linguistics}, pages 7828--7838, Association for Computational Linguistics, Online.

\bibitem[{Gandhi et~al.(2024)Gandhi, Fr{\"a}nken, Gerstenberg, and Goodman}]{gandhi2024understanding}
Gandhi, Kanishk, Jan-Philipp Fr{\"a}nken, Tobias Gerstenberg, and Noah Goodman. 2024.
\newblock Understanding social reasoning in language models with language models.
\newblock \emph{Advances in Neural Information Processing Systems}, 36.

\bibitem[{Gehman et~al.(2020)Gehman, Gururangan, Sap, Choi, and Smith}]{gehman2020realtoxicityprompts}
Gehman, Samuel, Suchin Gururangan, Maarten Sap, Yejin Choi, and Noah~A Smith. 2020.
\newblock Realtoxicityprompts: Evaluating neural toxic degeneration in language models.
\newblock \emph{arXiv preprint arXiv:2009.11462}.

\bibitem[{Gonen and Goldberg(2019)}]{gonen-goldberg-2019-lipstick}
Gonen, Hila and Yoav Goldberg. 2019.
\newblock Lipstick on a pig: {D}ebiasing methods cover up systematic gender biases in word embeddings but do not remove them.
\newblock In \emph{Proceedings of the 2019 Conference of the North {A}merican Chapter of the Association for Computational Linguistics: Human Language Technologies, Volume 1 (Long and Short Papers)}, pages 609--614, Association for Computational Linguistics, Minneapolis, Minnesota.

\bibitem[{Grant, Nematzadeh, and Griffiths(2017)}]{grant2017can}
Grant, Erin, Aida Nematzadeh, and Thomas~L Griffiths. 2017.
\newblock How can memory-augmented neural networks pass a false-belief task?
\newblock In \emph{CogSci}.

\bibitem[{Guo et~al.(2020)Guo, Kamar, Vaughan, Wallach, and Morris}]{guo2020toward}
Guo, Anhong, Ece Kamar, Jennifer~Wortman Vaughan, Hanna Wallach, and Meredith~Ringel Morris. 2020.
\newblock Toward fairness in ai for people with disabilities sbg@ a research roadmap.
\newblock \emph{ACM SIGACCESS Accessibility and Computing}, (125):1--1.

\bibitem[{Held et~al.(2023)Held, Harris, Best, and Yang}]{held2023material}
Held, William, Camille Harris, Michael Best, and Diyi Yang. 2023.
\newblock A material lens on coloniality in nlp.
\newblock \emph{arXiv preprint arXiv:2311.08391}.

\bibitem[{Hernez-Broome(2012)}]{hernez2012social}
Hernez-Broome, Gina. 2012.
\newblock Social intelligence: the new science of human relationships.

\bibitem[{Hershcovich et~al.(2022)Hershcovich, Frank, Lent, de~Lhoneux, Abdou, Brandl, Bugliarello, Cabello~Piqueras, Chalkidis, Cui, Fierro, Margatina, Rust, and S{\o}gaard}]{hershcovich-etal-2022-challenges}
Hershcovich, Daniel, Stella Frank, Heather Lent, Miryam de~Lhoneux, Mostafa Abdou, Stephanie Brandl, Emanuele Bugliarello, Laura Cabello~Piqueras, Ilias Chalkidis, Ruixiang Cui, Constanza Fierro, Katerina Margatina, Phillip Rust, and Anders S{\o}gaard. 2022.
\newblock Challenges and strategies in cross-cultural {NLP}.
\newblock In \emph{Proceedings of the 60th Annual Meeting of the Association for Computational Linguistics (Volume 1: Long Papers)}, pages 6997--7013, Association for Computational Linguistics, Dublin, Ireland.

\bibitem[{Hovy, Bianchi, and Fornaciari(2020)}]{hovy2020you}
Hovy, Dirk, Federico Bianchi, and Tommaso Fornaciari. 2020.
\newblock {``Y}ou sound just like your father{'' C}ommercial machine translation systems include stylistic biases.
\newblock In \emph{Proceedings of the 58th Annual Meeting of the Association for Computational Linguistics}, pages 1686--1690, Association for Computational Linguistics, Online.

\bibitem[{Hovy and Spruit(2016)}]{hovy-spruit-2016-social}
Hovy, Dirk and Shannon~L. Spruit. 2016.
\newblock The social impact of natural language processing.
\newblock In \emph{Proceedings of the 54th Annual Meeting of the Association for Computational Linguistics (Volume 2: Short Papers)}, pages 591--598, Association for Computational Linguistics, Berlin, Germany.

\bibitem[{Hovy and Yang(2021)}]{hovy2021importance}
Hovy, Dirk and Diyi Yang. 2021.
\newblock The importance of modeling social factors of language: Theory and practice.
\newblock In \emph{Proceedings of the 2021 Conference of the North American Chapter of the Association for Computational Linguistics: Human Language Technologies}, pages 588--602.

\bibitem[{Huang and Chang(2023)}]{huang2023towards}
Huang, Jie and Kevin Chen-Chuan Chang. 2023.
\newblock Towards reasoning in large language models: A survey.
\newblock In \emph{61st Annual Meeting of the Association for Computational Linguistics, ACL 2023}, pages 1049--1065, Association for Computational Linguistics (ACL).

\bibitem[{Ivey et~al.(2024)Ivey, Kumar, Liu, Shen, Rakshit, Raju, Zhang, Ananthasubramaniam, Kim, Yi et~al.}]{ivey2024real}
Ivey, Jonathan, Shivani Kumar, Jiayu Liu, Hua Shen, Sushrita Rakshit, Rohan Raju, Haotian Zhang, Aparna Ananthasubramaniam, Junghwan Kim, Bowen Yi, et~al. 2024.
\newblock Real or robotic? assessing whether llms accurately simulate qualities of human responses in dialogue.
\newblock \emph{arXiv preprint arXiv:2409.08330}.

\bibitem[{Iyyer et~al.(2016)Iyyer, Guha, Chaturvedi, Boyd-Graber, and Daum{\'e}~III}]{iyyer2016feuding}
Iyyer, Mohit, Anupam Guha, Snigdha Chaturvedi, Jordan Boyd-Graber, and Hal Daum{\'e}~III. 2016.
\newblock Feuding families and former friends: Unsupervised learning for dynamic fictional relationships.
\newblock In \emph{Proceedings of the 2016 Conference of the North American Chapter of the Association for Computational Linguistics: Human Language Technologies}, pages 1534--1544.

\bibitem[{Jakesch et~al.(2023)Jakesch, Bhat, Buschek, Zalmanson, and Naaman}]{jakesch2023co}
Jakesch, Maurice, Advait Bhat, Daniel Buschek, Lior Zalmanson, and Mor Naaman. 2023.
\newblock Co-writing with opinionated language models affects users’ views.
\newblock In \emph{Proceedings of the 2023 CHI conference on human factors in computing systems}, pages 1--15.

\bibitem[{Jin et~al.(2021)Jin, Chauhan, Tse, Sachan, and Mihalcea}]{jin2021good}
Jin, Zhijing, Geeticka Chauhan, Brian Tse, Mrinmaya Sachan, and Rada Mihalcea. 2021.
\newblock How good is nlp? a sober look at nlp tasks through the lens of social impact.
\newblock In \emph{Findings of the Association for Computational Linguistics: ACL-IJCNLP 2021}, pages 3099--3113.

\bibitem[{Joshi et~al.(2020)Joshi, Santy, Budhiraja, Bali, and Choudhury}]{joshi-etal-2020-state}
Joshi, Pratik, Sebastin Santy, Amar Budhiraja, Kalika Bali, and Monojit Choudhury. 2020.
\newblock The state and fate of linguistic diversity and inclusion in the {NLP} world.
\newblock In \emph{Proceedings of the 58th Annual Meeting of the Association for Computational Linguistics}, pages 6282--6293, Association for Computational Linguistics, Online.

\bibitem[{Kantharuban et~al.(2024)Kantharuban, Milbauer, Strubell, and Neubig}]{kantharuban2024stereotype}
Kantharuban, Anjali, Jeremiah Milbauer, Emma Strubell, and Graham Neubig. 2024.
\newblock Stereotype or personalization? user identity biases chatbot recommendations.
\newblock \emph{arXiv preprint arXiv:2410.05613}.

\bibitem[{Kruk, Ziems, and Yang(2023)}]{kruk2023impressions}
Kruk, Julia, Caleb Ziems, and Diyi Yang. 2023.
\newblock Impressions: Visual semiotics and aesthetic impact understanding.
\newblock In \emph{Proceedings of the 2023 Conference on Empirical Methods in Natural Language Processing}, pages 12273--12291.

\bibitem[{Kumar, AbuHashem, and Durumeric(2024)}]{kumar2024watch}
Kumar, Deepak, Yousef~Anees AbuHashem, and Zakir Durumeric. 2024.
\newblock Watch your language: Investigating content moderation with large language models.
\newblock In \emph{Proceedings of the International AAAI Conference on Web and Social Media}, volume~18, pages 865--878.

\bibitem[{Kwon et~al.(2023)Kwon, Hu, Myers, Karamcheti, Dragan, and Sadigh}]{kwon2023toward}
Kwon, Minae, Hengyuan Hu, Vivek Myers, Siddharth Karamcheti, Anca Dragan, and Dorsa Sadigh. 2023.
\newblock Toward grounded social reasoning.
\newblock \emph{arXiv preprint arXiv:2306.08651}.

\bibitem[{Lapinski and Rimal(2005)}]{lapinski2005explication}
Lapinski, Maria~Knight and Rajiv~N Rimal. 2005.
\newblock An explication of social norms.
\newblock \emph{Communication theory}, 15(2):127--147.

\bibitem[{Lauscher et~al.(2022)Lauscher, Bianchi, Bowman, and Hovy}]{lauscher2022socioprobe}
Lauscher, Anne, Federico Bianchi, Samuel Bowman, and Dirk Hovy. 2022.
\newblock Socioprobe: What, when, and where language models learn about sociodemographics.
\newblock \emph{arXiv preprint arXiv:2211.04281}.

\bibitem[{Le, Boureau, and Nickel(2019)}]{le2019revisiting}
Le, Matthew, Y-Lan Boureau, and Maximilian Nickel. 2019.
\newblock Revisiting the evaluation of theory of mind through question answering.
\newblock In \emph{Proceedings of the 2019 Conference on Empirical Methods in Natural Language Processing and the 9th International Joint Conference on Natural Language Processing (EMNLP-IJCNLP)}, pages 5872--5877.

\bibitem[{Litschko et~al.(2023)Litschko, M{\"u}ller-Eberstein, Van Der~Goot, Weber, and Plank}]{litschko2023establishing}
Litschko, Robert, Max M{\"u}ller-Eberstein, Rob Van Der~Goot, Leon Weber, and Barbara Plank. 2023.
\newblock Establishing trustworthiness: Rethinking tasks and model evaluation.
\newblock \emph{arXiv preprint arXiv:2310.05442}.

\bibitem[{Liu et~al.(2022)Liu, Mittal, Yang, and Bruckman}]{liu2022will}
Liu, Yihe, Anushk Mittal, Diyi Yang, and Amy Bruckman. 2022.
\newblock Will ai console me when i lose my pet? understanding perceptions of ai-mediated email writing.
\newblock In \emph{Proceedings of the 2022 CHI conference on human factors in computing systems}, pages 1--13.

\bibitem[{Ma et~al.(2024)Ma, Wang, Hu, Haensch, Hedderich, Plank, and Kreuter}]{ma-etal-2024-potential}
Ma, Bolei, Xinpeng Wang, Tiancheng Hu, Anna-Carolina Haensch, Michael~A. Hedderich, Barbara Plank, and Frauke Kreuter. 2024.
\newblock The potential and challenges of evaluating attitudes, opinions, and values in large language models.
\newblock In \emph{Findings of the Association for Computational Linguistics: EMNLP 2024}, pages 8783--8805, Association for Computational Linguistics, Miami, Florida, USA.

\bibitem[{McCarthy et~al.(2006)McCarthy, Minsky, Rochester, and Shannon}]{mccarthy2006proposal}
McCarthy, John, Marvin~L Minsky, Nathaniel Rochester, and Claude~E Shannon. 2006.
\newblock A proposal for the dartmouth summer research project on artificial intelligence, august 31, 1955.
\newblock \emph{AI magazine}, 27(4):12--12.

\bibitem[{Mittal et~al.(2020)Mittal, Guhan, Bhattacharya, Chandra, Bera, and Manocha}]{mittal2020emoticon}
Mittal, Trisha, Pooja Guhan, Uttaran Bhattacharya, Rohan Chandra, Aniket Bera, and Dinesh Manocha. 2020.
\newblock Emoticon: Context-aware multimodal emotion recognition using frege's principle.
\newblock In \emph{Proceedings of the IEEE/CVF Conference on Computer Vision and Pattern Recognition}, pages 14234--14243.

\bibitem[{Moravec(1988)}]{moravec1988mind}
Moravec, Hans. 1988.
\newblock \emph{Mind children: The future of robot and human intelligence}.
\newblock Harvard University Press.

\bibitem[{Nguyen et~al.(2016)Nguyen, Do{\u{g}}ru{\"o}z, Ros{\'e}, and de~Jong}]{nguyen-etal-2016-survey}
Nguyen, Dong, A.~Seza Do{\u{g}}ru{\"o}z, Carolyn~P. Ros{\'e}, and Franciska de~Jong. 2016.
\newblock Computational sociolinguistics: A {S}urvey.
\newblock \emph{Computational Linguistics}, 42(3):537--593.

\bibitem[{Nguyen, Rosseel, and Grieve(2021)}]{nguyen2021learning}
Nguyen, Dong, Laura Rosseel, and Jack Grieve. 2021.
\newblock On learning and representing social meaning in nlp: a sociolinguistic perspective.
\newblock In \emph{Proceedings of the 2021 Conference of the North American Chapter of the Association for Computational Linguistics: Human Language Technologies}, pages 603--612.

\bibitem[{Park et~al.(2024)Park, Zou, Shaw, Hill, Cai, Morris, Willer, Liang, and Bernstein}]{Park2024GenerativeAS}
Park, Joon~Sung, Carolyn~Q. Zou, Aaron Shaw, Benjamin~Mako Hill, Carrie~Jun Cai, Meredith~Ringel Morris, Robb Willer, Percy Liang, and Michael~S. Bernstein. 2024.
\newblock Generative agent simulations of 1,000 people.

\bibitem[{Pawar et~al.(2024)Pawar, Park, Jin, Arora, Myung, Yadav, Haznitrama, Song, Oh, and Augenstein}]{pawar2024surveyculturalawarenesslanguage}
Pawar, Siddhesh, Junyeong Park, Jiho Jin, Arnav Arora, Junho Myung, Srishti Yadav, Faiz~Ghifari Haznitrama, Inhwa Song, Alice Oh, and Isabelle Augenstein. 2024.
\newblock Survey of cultural awareness in language models: Text and beyond.

\bibitem[{Pentland(2005)}]{pentland2005socially}
Pentland, Alex. 2005.
\newblock Socially aware computation and communication.
\newblock In \emph{Proceedings of the 7th international conference on Multimodal interfaces}, pages 199--199.

\bibitem[{Pinker(2003)}]{pinker2003language}
Pinker, Steven. 2003.
\newblock \emph{The language instinct: How the mind creates language}.
\newblock Penguin uK.

\bibitem[{Prabhakaran, John, and Seligmann(2013)}]{prabhakaran-etal-2013-upper}
Prabhakaran, Vinodkumar, Ajita John, and Dor{\'e}e~D. Seligmann. 2013.
\newblock Who had the upper hand? ranking participants of interactions based on their relative power.
\newblock In \emph{Proceedings of the Sixth International Joint Conference on Natural Language Processing}, pages 365--373, Asian Federation of Natural Language Processing, Nagoya, Japan.

\bibitem[{Premack and Woodruff(1978)}]{premack1978does}
Premack, David and Guy Woodruff. 1978.
\newblock Does the chimpanzee have a theory of mind?
\newblock \emph{Behavioral and brain sciences}, 1(4):515--526.

\bibitem[{Ranathunga and de~Silva(2022)}]{ranathunga2022some}
Ranathunga, Surangika and Nisansa de~Silva. 2022.
\newblock Some languages are more equal than others: Probing deeper into the linguistic disparity in the nlp world.
\newblock \emph{arXiv preprint arXiv:2210.08523}.

\bibitem[{Rathje et~al.(2024)Rathje, Mirea, Sucholutsky, Marjieh, Robertson, and Van~Bavel}]{rathje2024gpt}
Rathje, Steve, Dan-Mircea Mirea, Ilia Sucholutsky, Raja Marjieh, Claire~E Robertson, and Jay~J Van~Bavel. 2024.
\newblock Gpt is an effective tool for multilingual psychological text analysis.
\newblock \emph{Proceedings of the National Academy of Sciences}, 121(34):e2308950121.

\bibitem[{Rhodes and Baron(2019)}]{rhodes2019development}
Rhodes, Marjorie and Andrew Baron. 2019.
\newblock The development of social categorization.
\newblock \emph{Annual review of developmental psychology}, 1(1):359--386.

\bibitem[{Ryan, Held, and Yang(2024)}]{ryan-etal-2024-unintended}
Ryan, Michael~J, William Held, and Diyi Yang. 2024.
\newblock Unintended impacts of {LLM} alignment on global representation.
\newblock In \emph{Proceedings of the 62nd Annual Meeting of the Association for Computational Linguistics (Volume 1: Long Papers)}, pages 16121--16140, Association for Computational Linguistics, Bangkok, Thailand.

\bibitem[{Sap et~al.(2022)Sap, Le~Bras, Fried, and Choi}]{sap-etal-2022-neural}
Sap, Maarten, Ronan Le~Bras, Daniel Fried, and Yejin Choi. 2022.
\newblock Neural theory-of-mind? on the limits of social intelligence in large {LM}s.
\newblock In \emph{Proceedings of the 2022 Conference on Empirical Methods in Natural Language Processing}, pages 3762--3780, Association for Computational Linguistics, Abu Dhabi, United Arab Emirates.

\bibitem[{Shah, Schwartz, and Hovy(2020)}]{shah-etal-2020-predictive}
Shah, Deven~Santosh, H.~Andrew Schwartz, and Dirk Hovy. 2020.
\newblock Predictive biases in natural language processing models: A conceptual framework and overview.
\newblock In \emph{Proceedings of the 58th Annual Meeting of the Association for Computational Linguistics}, pages 5248--5264, Association for Computational Linguistics, Online.

\bibitem[{Sharma et~al.(2023)Sharma, Rushton, Lin, Wadden, Lucas, Miner, Nguyen, and Althoff}]{sharma2023cognitive}
Sharma, Ashish, Kevin Rushton, Inna~Wanyin Lin, David Wadden, Khendra~G Lucas, Adam~S Miner, Theresa Nguyen, and Tim Althoff. 2023.
\newblock Cognitive reframing of negative thoughts through human-language model interaction.
\newblock \emph{arXiv preprint arXiv:2305.02466}.

\bibitem[{Sher, Koenig, and Rustichini(2014)}]{sher2014children}
Sher, Itai, Melissa Koenig, and Aldo Rustichini. 2014.
\newblock Children’s strategic theory of mind.
\newblock \emph{Proceedings of the National Academy of Sciences}, 111(37):13307--13312.

\bibitem[{Shu et~al.(2024)Shu, Zhang, Choi, Dunagan, Card, and Jurgens}]{shu2024you}
Shu, Bangzhao, Lechen Zhang, Minje Choi, Lavinia Dunagan, Dallas Card, and David Jurgens. 2024.
\newblock You don't need a personality test to know these models are unreliable: Assessing the reliability of large language models on psychometric instruments.
\newblock In \emph{Proceedings of the 2024 Annual Conference of the North American Chapter of the Association for Computational Linguistics}.

\bibitem[{Snyder and Ickes(1985)}]{snyder1985personality}
Snyder, Mark and William Ickes. 1985.
\newblock Personality and social behavior.
\newblock \emph{Handbook of social psychology}, 2(3):883--947.

\bibitem[{Song et~al.(2023)Song, Cui, Khanuja, Liu, Faisal, Ostapenko, Winata, Aji, Cahyawijaya, Tsvetkov et~al.}]{song2023globalbench}
Song, Yueqi, Catherine Cui, Simran Khanuja, Pengfei Liu, Fahim Faisal, Alissa Ostapenko, Genta~Indra Winata, Alham~Fikri Aji, Samuel Cahyawijaya, Yulia Tsvetkov, et~al. 2023.
\newblock Globalbench: A benchmark for global progress in natural language processing.
\newblock \emph{arXiv preprint arXiv:2305.14716}.

\bibitem[{Soni et~al.(2024)Soni, Flek, Sharma, Yang, Hooker, and Schwartz}]{hucllm-2024-human}
Soni, Nikita, Lucie Flek, Ashish Sharma, Diyi Yang, Sara Hooker, and H.~Andrew Schwartz, editors. 2024.
\newblock \emph{Proceedings of the 1st Human-Centered Large Language Modeling Workshop}. ACL, TBD.

\bibitem[{Sorensen et~al.(2024)Sorensen, Moore, Fisher, Gordon, Mireshghallah, Rytting, Ye, Jiang, Lu, Dziri et~al.}]{sorensen2024roadmap}
Sorensen, Taylor, Jared Moore, Jillian Fisher, Mitchell Gordon, Niloofar Mireshghallah, Christopher~Michael Rytting, Andre Ye, Liwei Jiang, Ximing Lu, Nouha Dziri, et~al. 2024.
\newblock A roadmap to pluralistic alignment.
\newblock \emph{arXiv preprint arXiv:2402.05070}.

\bibitem[{Spangher et~al.(2024)Spangher, Peng, Gehrmann, and Dredze}]{spangher-etal-2024-llms}
Spangher, Alexander, Nanyun Peng, Sebastian Gehrmann, and Mark Dredze. 2024.
\newblock Do {LLM}s plan like human writers? comparing journalist coverage of press releases with {LLM}s.
\newblock In \emph{Proceedings of the 2024 Conference on Empirical Methods in Natural Language Processing}, pages 21814--21828, Association for Computational Linguistics, Miami, Florida, USA.

\bibitem[{Suh, Althoff, and Torous(2024)}]{suh2024special}
Suh, Jina, Tim Althoff, and John Torous. 2024.
\newblock Special report: Are you ready for generative ai in psychiatric practice?

\bibitem[{Tedeschi et~al.(2023)Tedeschi, Bos, Declerck, Haji{\v{c}}, Hershcovich, Hovy, Koller, Krek, Schockaert, Sennrich, Shutova, and Navigli}]{tedeschi-etal-2023-whats}
Tedeschi, Simone, Johan Bos, Thierry Declerck, Jan Haji{\v{c}}, Daniel Hershcovich, Eduard Hovy, Alexander Koller, Simon Krek, Steven Schockaert, Rico Sennrich, Ekaterina Shutova, and Roberto Navigli. 2023.
\newblock What{'}s the meaning of superhuman performance in today{'}s {NLU}?
\newblock In \emph{Proceedings of the 61st Annual Meeting of the Association for Computational Linguistics (Volume 1: Long Papers)}, pages 12471--12491, Association for Computational Linguistics, Toronto, Canada.

\bibitem[{Thompson(2007)}]{thompson2007development}
Thompson, Ross~A. 2007.
\newblock The development of the person: Social understanding, relationships, conscience, self.
\newblock \emph{Handbook of Child Psychology}, 3.

\bibitem[{Tian et~al.(2024)Tian, Huang, Liu, Jiang, Spangher, Chen, May, and Peng}]{tian-etal-2024-large-language}
Tian, Yufei, Tenghao Huang, Miri Liu, Derek Jiang, Alexander Spangher, Muhao Chen, Jonathan May, and Nanyun Peng. 2024.
\newblock Are large language models capable of generating human-level narratives?
\newblock In \emph{Proceedings of the 2024 Conference on Empirical Methods in Natural Language Processing}, pages 17659--17681, Association for Computational Linguistics, Miami, Florida, USA.

\bibitem[{Tomasello(2014)}]{tomasello2014natural}
Tomasello, Michael. 2014.
\newblock \emph{A natural history of human thinking}.
\newblock Harvard University Press.

\bibitem[{Turing(1950)}]{turing1950}
Turing, Alan~M. 1950.
\newblock {I.—COMPUTING MACHINERY AND INTELLIGENCE}.
\newblock \emph{Mind}, LIX(236):433--460.

\bibitem[{Vanmassenhove, Hardmeier, and Way(2018)}]{vanmassenhove-etal-2018-getting}
Vanmassenhove, Eva, Christian Hardmeier, and Andy Way. 2018.
\newblock Getting gender right in neural machine translation.
\newblock In \emph{Proceedings of the 2018 Conference on Empirical Methods in Natural Language Processing}, pages 3003--3008, Association for Computational Linguistics, Brussels, Belgium.

\bibitem[{Vida, Damken, and Lauscher(2024)}]{vida2024decoding}
Vida, Karina, Fabian Damken, and Anne Lauscher. 2024.
\newblock Decoding multilingual moral preferences: Unveiling llm's biases through the moral machine experiment.
\newblock In \emph{Proceedings of the AAAI/ACM Conference on AI, Ethics, and Society}, volume~7, pages 1490--1501.

\bibitem[{Wang et~al.(2024)Wang, Zhang, Robinson, Loeb, and Demszky}]{wang-etal-2024-bridging}
Wang, Rose~E., Qingyang Zhang, Carly Robinson, Susanna Loeb, and Dorottya Demszky. 2024.
\newblock Bridging the novice-expert gap via models of decision-making: A case study on remediating math mistakes.
\newblock In \emph{Proceedings of the 2024 Conference of the North American Chapter of the Association for Computational Linguistics}, Association for Computational Linguistics.

\bibitem[{Wilson and Caliskan(2024)}]{wilson2024gender}
Wilson, Kyra and Aylin Caliskan. 2024.
\newblock Gender, race, and intersectional bias in resume screening via language model retrieval.
\newblock In \emph{Proceedings of the AAAI/ACM Conference on AI, Ethics, and Society}, volume~7, pages 1578--1590.

\bibitem[{Wright et~al.(2024)Wright, Arora, Borenstein, Yadav, Belongie, and Augenstein}]{wright-etal-2024-llm}
Wright, Dustin, Arnav Arora, Nadav Borenstein, Srishti Yadav, Serge Belongie, and Isabelle Augenstein. 2024.
\newblock {LLM} tropes: Revealing fine-grained values and opinions in large language models.
\newblock In \emph{Findings of the Association for Computational Linguistics: EMNLP 2024}, pages 17085--17112, Association for Computational Linguistics, Miami, Florida, USA.

\bibitem[{Yin et~al.(2021)Yin, Moryossef, Hochgesang, Goldberg, and Alikhani}]{yin2021including}
Yin, Kayo, Amit Moryossef, Julie Hochgesang, Yoav Goldberg, and Malihe Alikhani. 2021.
\newblock Including signed languages in natural language processing.
\newblock \emph{arXiv preprint arXiv:2105.05222}.

\bibitem[{Zhang et~al.(2023)Zhang, Xu, Zhang, Liu, Hooi, and Deng}]{zhang2023exploring}
Zhang, Jintian, Xin Xu, Ningyu Zhang, Ruibo Liu, Bryan Hooi, and Shumin Deng. 2023.
\newblock Exploring collaboration mechanisms for llm agents: A social psychology view.
\newblock \emph{arXiv preprint arXiv:2310.02124}.

\bibitem[{Ziems et~al.(2023)Ziems, Shaikh, Zhang, Held, Chen, and Yang}]{ziems2023can}
Ziems, Caleb, Omar Shaikh, Zhehao Zhang, William Held, Jiaao Chen, and Diyi Yang. 2023.
\newblock Can large language models transform computational social science?
\newblock \emph{Computational Linguistics}, pages 1--53.

\end{thebibliography}

\end{document}